\documentclass[11pt, a4paper, logo, copyright, nonumbering]{main}
\usepackage[authoryear, sort&compress, round]{natbib}
\usepackage{dblfloatfix}
\usepackage{ulem}
\usepackage{caption}
\usepackage{dramatist}
\usepackage{xspace}
\usepackage{pifont} %
\usepackage{multirow}
\usepackage{tcolorbox}
\usepackage{xltabular}
\usepackage{longtable}
\usepackage{hyperref}

\usepackage{amsfonts}
\usepackage{amsmath}
\usepackage{amssymb}
\usepackage{lineno}
\usepackage{multirow}
\usepackage{adjustbox}

\usepackage[bottom]{footmisc}
\usepackage{subcaption}
\usepackage{graphicx}
\usepackage{CJKutf8}
\usepackage{setspace}

\usepackage{dsfont}
\usepackage{array} %
\usepackage{tabularx} %
\usepackage{xcolor} %
\usepackage{tabularx}
\usepackage{booktabs}
\usepackage{xspace}

\usepackage{lipsum}  %
\usepackage{multicol} %

\usepackage{cleveref}
\usepackage{epigraph}
\usepackage{nicematrix}
\usepackage{appendix}
\usepackage{hyperref}      
\usepackage{fontawesome5}  
\usepackage{xcolor}        
\usepackage{tcolorbox}
\tcbuselibrary{breakable}
\tcbuselibrary{skins}
\usepackage{wrapfig}
\definecolor{groupgray}{gray}{0.9}
\newcommand{\methodname}{\textsc{Prime}\xspace}

\makeatletter
\def\@BTrule[#1]{%
  \ifx\longtable\undefined
    \let\@BTswitch\@BTnormal
  \else\ifx\hline\LT@hline
    \nobreak
    \let\@BTswitch\@BLTrule
  \else
    \let\@BTswitch\@BTnormal
  \fi\fi
  \global\@thisrulewidth=#1\relax
  \ifnum\@thisruleclass=\tw@\vskip\@aboverulesep\else
  \ifnum\@lastruleclass=\z@\vskip\@aboverulesep\else
  \ifnum\@lastruleclass=\@ne\vskip\doublerulesep\fi\fi\fi
  \@BTswitch
}
\makeatother

\addto\extrasenglish{
}

{\begin{list}{}%
        {\setlength{\leftmargin}{#1}}%
        \item[]%
}
{\end{list}}

\reportnumber{001} %

\renewcommand{\today}{}

\title{PRIME: A Process-Outcome Alignment Benchmark for Verifiable Reasoning in Mathematics and Engineering}

\def\huggingface{\raisebox{-1.5pt}{\includegraphics[height=1.05em]{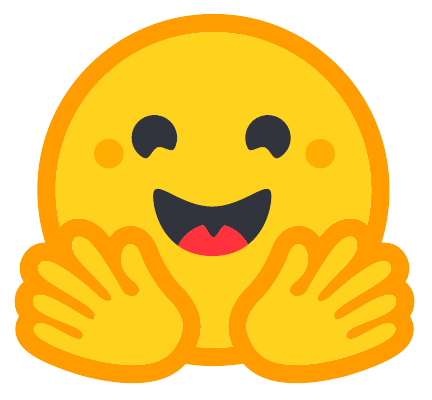}}}
\def\github{\raisebox{-1.5pt}{\includegraphics[height=1.05em]{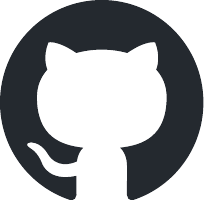}}}

\author[*]{
    \centering
    \parbox{0.9\textwidth}{\centering
    \large \strut \textbf{Xiangfeng Wang}$^{1,2,*}$, \textbf{Hangyu Guo}$^{2,*}$, \textbf{Yanlin Lai}$^{2,3,*}$, \textbf{Mitt Huang}$^{2,*}$, \quad \quad
    \large \strut \textbf{Liang Zhao}$^{2}$, \textbf{Chengyuan Yao}$^{2}$, \textbf{Yinmin Zhang}$^{2}$, \textbf{Qi Han}$^{2}$, \textbf{Xiaoxiao Ren}$^{2}$, \quad \quad
    \large \strut \textbf{Chun Yuan}$^{3,\dagger}$, \textbf{Tong Xu}$^{1,\dagger}$, \textbf{Zheng Ge}$^{2}$, \textbf{Xiangyu Zhang}$^{2}$, \textbf{Daxin Jiang}$^{2}$ \strut
    }
    
    {\small
        \normalfont
        $^{1}$University of Science and Technology of China \quad $^{2}$StepFun \quad $^{3}$Tsinghua University
        \\
        {\small \normalfont{$^{*}$Equal contribution \quad $^{\dagger}$Corresponding Author}}
    }
    \vspace{0.5em}
    
    \small
    \github~\textbf{Github}: \url{https://github.com/wonderful9462/PRIME} \\
    \small
    \huggingface~\textbf{Huggingface}: \href{https://huggingface.co/collections/wonderful9462/prime}{PRIME Collections}
}

\newcommand{\model}{\textsc{Model}}

\renewcommand{\phi}{\varphi}

\renewcommand{\epsilon}{\varepsilon}
\renewcommand{\imath}{\mathrm{i}}

\newlength{\restsubwidth}
\newlength{\restsubheight}
\newlength{\restsubmoreheight}
\newcommand{\rest}[2]{%
        \settowidth{\restsubwidth}{\ensuremath{#2}}
        \settoheight{\restsubheight}{\ensuremath{{}_{#2}}}
        \ensuremath{{#1\hskip 0.5pt}_{\vrule\kern2pt\parbox[b][%
        4pt][b]{\the\restsubwidth}{%
                        \ensuremath{{}_{#2}}}}}
        }

\begin{abstract}

While model-based verifiers are essential for scaling Reinforcement Learning with Verifiable Rewards (RLVR), current outcome-centric verification paradigms primarily focus on the consistency between the final result and the ground truth, often neglecting potential errors in the derivation process. This leads to assigning positive rewards to correct answers produced from incorrect derivations. To bridge this gap, we introduce \textbf{\methodname}, a benchmark for evaluating verifiers on \textbf{P}\textbf{R}ocess-outcome alignment verification \textbf{I}n \textbf{M}athematics and \textbf{E}ngineering. Curated from a comprehensive collection of college-level STEM problems, \methodname comprises 2,530 high-difficulty samples through a consistency-based filtering pipeline. Through extensive evaluation, we find that current verifiers frequently fail to detect derivation flaws. Furthermore, we propose a process-aware RLVR training paradigm utilizing verifiers selected via \methodname. This approach substantially outperforms the outcome-only verification baseline, achieving absolute performance gains of \textbf{8.29\%}, \textbf{9.12\%}, and \textbf{7.31\%} on AIME24, AIME25, and Beyond-AIME, respectively, for the Qwen3-14B-Base model. Finally, we demonstrate a strong linear correlation ($R^2 > 0.92$) between verifier accuracy on \methodname and RLVR training effectiveness, validating \methodname as a reliable predictor for verifier selection. We release the benchmark and code at \url{https://github.com/wonderful9462/PRIME}.
\end{abstract}

\begin{document}

\maketitle


\definecolor{colorfirst}{RGB}{252,141,89}
\definecolor{colorsecond}{RGB}{253,187,132}
\definecolor{colorthird}{RGB}{253,212,158}
\definecolor{colorfourth}{RGB}{254,232,200}
\definecolor{colorfifth}{RGB}{255,247,236}
\definecolor{myred}{RGB}{242,128,128}
\definecolor{mygreen}{RGB}{112,180,143}
\definecolor{myblue}{RGB}{210,225,255}
\definecolor{citypink}{RGB}{227,108,194}
\definecolor{cityblue}{RGB}{128,159,225}
\newcommand{\ph}[1]{\textcolor{black}{#1}}
\newcommand{\rankfirst}[0]{\cellcolor{colorfirst}}
\newcommand{\ranksecond}[0]{\cellcolor{colorsecond}}
\newcommand{\rankthird}[0]{\cellcolor{colorthird}}
\newcommand{\rankfourth}[0]{\cellcolor{colorfourth}}
\newcommand{\rankfifth}[0]{\cellcolor{colorfifth}}
\DeclareRobustCommand{\legendsquare}[1]{%
  \textcolor{#1}{\rule{2ex}{2ex}}%
}
\DeclareRobustCommand{\legendsquarebox}[1]{%
  \tikz[] \draw[black, fill=#1, line width=0.4pt] (0,0) rectangle (1.5ex,1.5ex);%
}
\newcommand{\cmark}{\textcolor{mygreen}{\ding{51}}}%
\newcommand{\xmark}{\textcolor{myred}{\ding{55}}}%



\section{Introduction}

Reinforcement learning with verifiable reward (RLVR) has proven highly effective for Large Reasoning Models (LRMs) in STEM domains~\citep{shao2024deepseekmath, guo2025deepseek}.
Given that rule-based verification frameworks, exemplified by Math-Verify\footnote{\url{https://github.com/huggingface/Math-Verify}}, often struggle with flexible output formats~\citep{wang2024math}, the field has transitioned toward model-based verifiers to provide more robust reward signals~\citep{liu2025deepseek, team2025kimi, liu-compassverify-2025-arxiv}.
However, it has become increasingly evident that the efficacy of RLVR is strictly bounded by the verifier's own capability, as variability in verification precision directly dictates the quality of the training outcome~\citep{mahan2024generative, zhang2025survey}. In response to this dependency, recent research has concentrated on rigorously benchmarking and optimizing verifier performance. Benchmarks such as 
\begin{wrapfigure}{r}{0.5\textwidth}
    \centering
    \includegraphics[width=\linewidth]{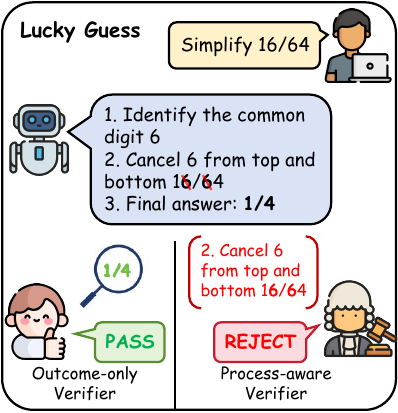}
    \captionsetup{singlelinecheck=false, justification=raggedright}
    \caption{\textbf{An illustration of a ``lucky guess''} where the model arrives at the correct answer via an incorrect derivation.}
    \label{fig: luckyguess}
    \vspace{-6mm}
\end{wrapfigure}
VerifierBench~\citep{liu-compassverify-2025-arxiv} and VAR~\citep{chen-xverify-arxiv-2025} have been established to evaluate judgment reliability, guiding the development of specialized methods~\citep{zheng-sciverifier-arxiv-2025, feng2025cosineverifier} aimed at precise verification across complex STEM tasks.

However, current verification benchmarks primarily focus on outcome correctness while neglecting the derivation process. As illustrated in Figure~\ref{fig: luckyguess}, this outcome-only approach erroneously validates ``lucky guesses,'' where the model arrives at the correct answer through incorrect derivation. To quantify this, we collect 2,500 STEM prompts and generate multiple responses using diverse open-source and closed-source models. Through rigorous human annotation, we identify that 16.98\% of the responses exhibit such ``lucky guesses,'' where the outcome is correct, but the derivation is flawed. Given this high frequency, verifiers guided by existing outcome-level verification benchmarks provide inaccurate rewards that reinforce flawed derivation, potentially limiting the performance of RLVR training.

To bridge this gap, we introduce \textbf{\methodname}, a benchmark for \textbf{P}\textbf{R}ocess–outcome alignment verification \textbf{I}n \textbf{M}athematics and \textbf{E}ngineering domains. Unlike evaluations limited solely to the outcome level, \methodname assesses verifiers' capability to detect both outcome accuracy and process-outcome logical consistency. To ensure the benchmark's reliability, we first collect a large pool of college-level STEM questions and apply model-based filtering to retain only verifiable problems with reliable reference answers. Second, we generate multiple solution trajectories for each retained problem using a diverse set of open source and commercial LRMs to cover a broad range of reasoning behaviors. Third, we perform difficulty-oriented selection to prioritize cases that are hard to verify. Finally, to guarantee the reliability of the dataset, we employ 18 domain experts to rigorously annotate each sample for both process-outcome consistency and outcome-ground truth alignment. The resulting benchmark comprises approximately 2,530 annotated instances, spanning 16 major STEM categories with fine-grained coverage across 480 subdisciplines.

In summary, our main contributions are threefold:
\begin{itemize}
\item We introduce \textbf{\methodname}, a high-quality benchmark specifically designed to evaluate process-outcome alignment in Mathematics and Engineering. Comprising 2,530 expert-annotated samples, it effectively detects spurious correctness and fills the gap in existing benchmarks that neglect derivation correctness.

\item We propose a process-aware RLVR training paradigm where employing a process-aware verifier grants rewards only when the LRM's final answer matches the ground truth and the derivation maintains logical consistency with its outcome. Compared to the outcome-only verifier, our approach achieves substantial performance gains, with absolute improvements of \textbf{8.29\%}, \textbf{9.12\%}, and \textbf{7.31\%} on AIME24, AIME25, and Beyond-AIME, respectively, on Qwen3-14B-Base model.

\item We demonstrate a strong linear correlation (with $R^2 > 0.92$) between a verifier's accuracy on \methodname\ and the resulting performance improvement in RLVR training. This validates \methodname\ as a reliable predictor for assessing and selecting effective verifiers for LRMs training.
\end{itemize}

\section{Related Works}

\subsection{Large Reasoning Models}
Recent advancements in Large Reasoning Models (LRMs) have been significantly propelled by Reinforcement Learning with Verifiable Rewards (RLVR), which incentivizes structured, multi-step problem-solving capabilities. Currently, mainstream RLVR paradigms primarily focus on outcome-centric verification, where the reward signal is determined solely by the consistency between the final result and the ground truth. Within this framework, verification methods are generally categorized into rule-based and model-based approaches. Rule-based approaches predominantly rely on deterministic systems, such as compilers for code or symbolic solvers for mathematics, to generate rigorous reward signals. Pioneering models like DeepSeek-R1~\citep{guo2025deepseek} and Qwen2.5-Math~\citep{yang2024qwen2} exemplify this direction, utilizing strict answer matching to drive large-scale reinforcement learning. To address the limitations of rigid string matching in flexible scenarios, \model-based approaches employ learned verifiers to assess the semantic equivalence between the model's final answer and the ground truth~\citep{seed2025seed1, zhang2024generative}. However, regardless of whether they are rule-based or model-based, these outcome-centric verifiers share a critical vulnerability: they are unable to detect inconsistencies in the intermediate reasoning process. This limitation prevents the effective suppression of the ``lucky guess'' phenomenon, where models receive positive reinforcement for correct answers derived from flawed logic.

\subsection{Verifier and Verifier Evaluation}
Recent advancements in model-based verification can be broadly categorized into outcome-centric and process-centric approaches. Outcome-centric methods, such as CompassVerifier~\citep{liu-compassverify-2025-arxiv} and xVerify~\citep{chen-xverify-arxiv-2025}, focus on assessing the semantic equivalence between generated answers and ground truth, with extensions like SCI-Verifier~\citep{zheng-sciverifier-arxiv-2025} further handling symbolic domains. However, by prioritizing the final result, these approaches often overlook the underlying derivation, failing to detect spurious correctness arising from ``lucky guesses.'' In contrast, process-centric approaches, known as Process Reward Models (PRMs)~\citep{uesato2022solving, lightman2023lets}, provide step-aware supervision to mitigate reasoning fallacies, with techniques like Math-Shepherd~\citep{wang2024math} automating verification via Monte Carlo rollouts. Despite the evolution of these verification paradigms, the evaluation of verifiers remains underdeveloped. Current benchmarks primarily test answer extraction capabilities rather than the ability to detect logical inconsistencies between the derivation and the final result. \methodname bridges this critical gap by introducing a benchmark for Process-Outcome Alignment, specifically penalizing reasoning false positives to distinguish valid reasoning from spurious success. \methodname addresses this limitation by introducing a specialized benchmark for Process-Outcome Alignment, which rigorously evaluates whether a verifier grants rewards only when the final answer matches the ground truth and the derivation maintains logical consistency with its outcome.

\section{Benchmark Construction}
\label{sec:construction}

\begin{figure*}[t]
    \centering
    \includegraphics[width=\textwidth]{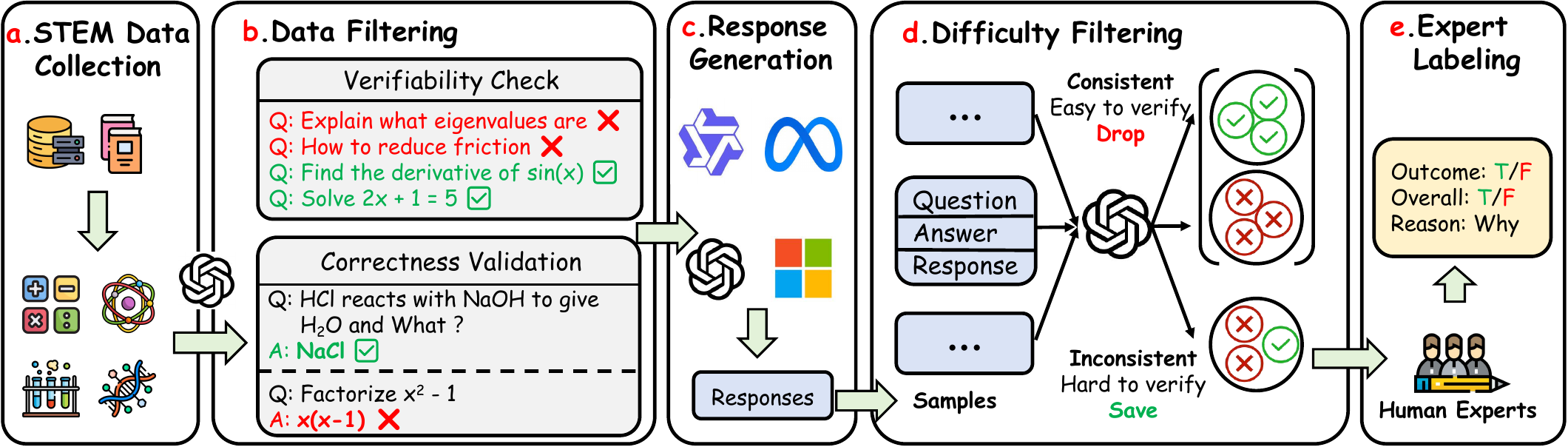}
    \captionsetup{singlelinecheck=false, justification=raggedright}
    \caption{Overview of the \textsc{Prime} construction pipeline. The process comprises five stages: (a) Extensive STEM data collection with diversity control; (b) Two-stage automated filtering for verifiability and correctness; (c) Heterogeneous response generation using diverse LRMs; (d) Difficulty-aware filtering to select discriminative samples; (e) Fine-grained expert labeling focusing on process-outcome alignment.}
    \label{fig:pipeline}
\end{figure*}

To systematically evaluate the capability of verifiers to judge the process-outcome consistency of LRMs, we introduce \textsc{Prime}.  The benchmark is constructed through a multi-stage pipeline that integrates automated processing with human-in-the-loop verification. As illustrated in Figure~\ref{fig:pipeline}, the construction encompasses five key stages, and we detail each one in the subsequent sections.

\subsection{STEM Data Collection and Filtering}
\label{sec:data_process}

To comprehensively evaluate the verification capability of large reasoning models across complex scientific domains, we construct an initial pool $\mathcal{Q}_{\text{raw}}$ containing over 7M problems collected primarily from undergraduate and graduate-level textbooks, as well as university final exams. These problems span Mathematics, Physics, Chemistry, and Engineering, requiring deep domain knowledge and complex multi-step reasoning. Raw data from educational resources often contains open-ended inquiries or errors unsuitable for objective verification. We further apply a two-stage model filtering pipeline to sanitize $\mathcal{Q}_{\text{raw}}$:

\begin{enumerate}
    \item \textbf{Verifiability Check:} We prompt GPT-OSS-120B to directly assess the problem statement for verifiability. This step filters out ambiguous or open-ended questions lacking a unique ground truth.
    \item \textbf{Correctness Validation:} To maximally eliminate errors in textbook answers, we employ top-performing models, including Gemini-3-Pro, GPT-5, and Claude-Sonnet-4.5, to cross-validate the ground truth $a_{\text{gt}}$. We filter out instances where none of the models yield a result matching $a_{\text{gt}}$, retaining only those validated by at least one model.
\end{enumerate}

\begin{wrapfigure}{r}{0.4\textwidth}
    \vspace{-2.2\baselineskip} 
    \centering
    \includegraphics[width=\linewidth]{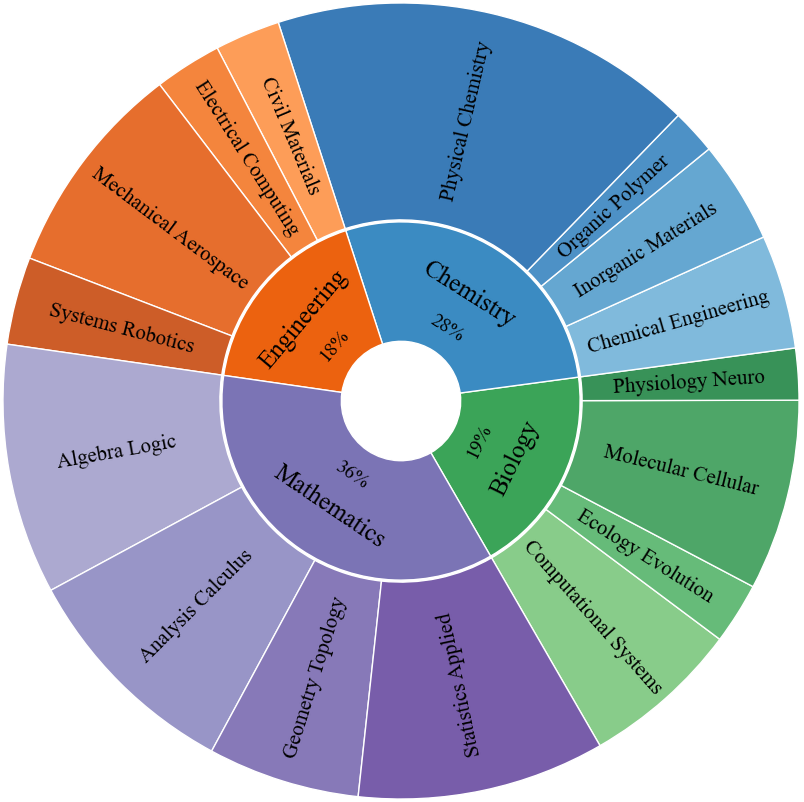}
    \captionsetup{singlelinecheck=false, justification=raggedright}
    \caption{\textbf{Subject distribution of \methodname.} The inner ring represents the four major STEM disciplines, while the outer ring details the 16 fine-grained sub-domains.}
    \label{fig:subject_dist}
\end{wrapfigure}

After filtering, we organize valid problems into a hierarchical structure encompassing 16 fine-grained sub-domains, such as Topology, Organic Polymer, and Systems Robotics. To prevent data imbalance caused by the uneven distribution of source materials, we apply a downsampling strategy to over-represented categories. We denote this final filtered dataset as $\mathcal{Q}_{\text{clean}} = \{(q_i, a_i)\}$, where $q_i$ represents the question statement and $a_i$ denotes the standard ground truth answer (typically a final value or option without reasoning steps). The subject composition is presented in Figure~\ref{fig:subject_dist}.

\subsection{Diverse Response Generation and \\ Difficulty-Aware Selection}
\label{sec:gen_and_filter}

To assess verifier robustness against diverse reasoning patterns, we construct a response set containing varying qualities and error types. For each valid problem $(q, a) \in \mathcal{Q}_{\text{clean}}$, we employ $M$ distinct Large Reasoning Models (LRMs) to serve as candidate solvers. Each model $\mathcal{M}_m$ generates a single reasoning trajectory $r_m$, yielding a set of $M$ triplets per question:
\begin{equation}
\mathcal{T} = \{ \left( q_i, a_i, {r_i^m}\right) |m \in \{M\}, (q_i, a_i)\in Q_{\text{clean}}\}
\end{equation}
where $r_i^m$ represents the response generated by the $m$-th model.

To ensure the benchmark focuses on samples that effectively discriminate verifier capabilities, we implement a \textit{Verification Difficulty Filtering} mechanism. We select responses that are intrinsically hard to verify. We utilize GPT-OSS-120B as a proxy verifier to assess each triplet $(q, a, r) \in \mathcal{T}$. The verifier is prompted to check whether the response $r$ is correct regarding both the reasoning process and the final result.

To capture the uncertainty in verification, we perform $k=8$ independent verification trials for each triplet. Let $v_j \in \{0, 1\}$ denote the verdict of the $j$-th trial, where $1$ indicates the response is judged as correct and $0$ otherwise. We define the \textit{Verification Consensus Score} $C(q, r)$ as the average of positive judgments: $C(q, r) = \sum_{j=1}^k v_j/k$. Based on this score, we categorize the samples:

\begin{itemize}
    \item \textbf{Easy-to-Verify (Drop):} $C(q, r) = 1$ or $C(q, r) = 0$. These are samples where the proxy verifier is unanimous across all trials (either consistently correct or consistently incorrect). These cases represent "obvious" decisions that offer minimal signal for distinguishing strong verifiers.
    \item \textbf{Hard-to-Verify (Save):} $0 < C(q, r) < 1$. These samples lie on the decision boundary of the verifier, where the proxy model exhibits inconsistency in its judgment. Such instances typically contain subtle errors or deceptive steps that are challenging to verify reliably.
\end{itemize}

We retain only the response triplets from the ``Hard-to-Verify'' category to form the final evaluation set $\mathcal{T}_{\text{final}}$. This strategy ensures that \textsc{Prime} specifically targets the difficult regime of process supervision.

\subsection{Fine-grained Expert Labeling and Evaluation Metrics}

To ensure the accuracy and reliability of our \methodname, we employ domain experts to perform fine-grained annotation on the selected samples in $\mathcal{T}_{\text{final}}$. Instead of relying on traditional outcome-only checks, we introduce a process-aware labeling. Specifically, for each trajectory $(q, r, a)$, experts provide a structured label tuple $L = (y_{\text{outcome}}, y_{\text{overall}})$:

\begin{itemize}
    \item $y_{\text{outcome}} \in \{0, 1\}$: \textbf{Outcome Label.} Validity of the extracted final answer $a$ against the ground truth $a_{\text{gt}}$.
    \item $y_{\text{overall}} \in \{0, 1\}$: \textbf{Overall Label (Process-Outcome Consistency).} This is the primary metric of \methodname. It is marked as valid ($1$) \textit{if and only if} the reasoning process $s$ is logically sound \textit{and} correctly derives $a$. This strictly penalizes ``lucky guesses'', where the answer is correct by chance but the derivation is flawed.
\end{itemize}

Consequently, the evaluation score on \methodname is calculated based on the prediction of the overall Label ($y_{\text{overall}}$), rather than the outcome-only label. This ensures that verifiers are assessed on their ability to validate the entire derivation process and final answer. The detailed prompts used for the verifier in our evaluation are provided in Appendix Figure~\ref{fig:eval_prompt}.

\section{Analysis of Performance on \methodname}
\label{bmk_result}

\begin{table*}[t]
\centering
\resizebox{\textwidth}{!}{
\begin{NiceTabular}{l |cc cc cc cc|ccc}
\toprule
\multirow{2}{*}{Model} 
& \multicolumn{2}{c}{Math} 
& \multicolumn{2}{c}{Physics} 
& \multicolumn{2}{c}{Chemistry} 
& \multicolumn{2}{c}{Biology} 
& \multicolumn{3}{c}{Average} \\
\cmidrule(lr){2-3}
\cmidrule(lr){4-5}
\cmidrule(lr){6-7}
\cmidrule(lr){8-9}
\cmidrule(lr){10-12}
& Outcome & Overall
& Outcome & Overall
& Outcome & Overall
& Outcome & Overall
& Outcome & Overall & F1 \\
\midrule
\rowcolor{groupgray}
\Block{1-11}{\textbf{Rule-based Verifier}}& & & & & & & & & & \\
\midrule
Math-Verify & 65.56 & 63.67 & 63.17 & 65.85 & 69.69 & 65.47 & 73.80 & 72.04 & 67.71 & 65.93 & 57.88 \\
\midrule
\rowcolor{groupgray}
\Block{1-11}{\textbf{Open-source Models}}& & & & & & & & & & \\
\midrule
Qwen3-4B-Instruct-2507 & 81.22 & 64.78 & 78.35 & 68.97 & 80.05 & 70.20 & 77.33 & 74.56 & 79.74 & 73.13 & 68.74 \\
Phi-4 & 81.44 & 60.00 & 81.70 & 70.09 & 80.05 & 63.81 & 76.57 & 69.52 & 80.29 & 64.46 & 69.08 \\
DeepSeek-V3.2 & 87.56 & 72.44 & 87.50 & 79.02 & 88.24 & 77.62 & 87.91 & 84.63 & 87.81 & 77.13 & 76.33 \\
\midrule
Qwen3-4B & 86.33 & 69.56 & 87.72 & 79.46 & 86.57 & 75.83 & 84.89 & 82.12 & 86.43 & 75.23 & 74.26 \\
Qwen3-4B-thinking-2507 & 89.22 & 76.22 & 89.51 & 82.81 & 91.30 & 83.76 & \underline{90.68} & 86.65 & \underline{90.15} & 81.36 & 79.88\\
Qwen3-8B & 87.78 & 72.89 & 88.17 & 80.13 & 87.47 & 77.37 & 87.91 & 84.13 & 87.77 & 77.32 & 76.62 \\
Qwen3-30B-A3B & 90.44 & 79.67 & 87.28 & 83.26 & 88.24 & 83.38 & 90.18 & \underline{87.66} & 89.16 & 82.71 & 80.34\\
GPT-OSS-20B & 89.67 & 75.56 & 88.62 & 83.26 & 88.24 & 79.80 & 89.67 & 85.14 & 89.04 & 79.74 & 78.75\\
GPT-OSS-120B & \underline{91.22} & 78.33 & 88.39 & 82.37 & 90.03 & 79.80 & 88.92 & 85.14 & 89.99 & 81.64 & 80.71\\
DeepSeek-V3.2-thinking & 90.67 & \textbf{83.89} & \underline{89.73} & \textbf{87.72} & \textbf{93.35} & \underline{88.49} & \textbf{91.18} & \textbf{90.43} & \textbf{91.41} & \textbf{87.02} & \textbf{85.65} \\
GLM-4.6 & 90.22 & 81.00 & 89.51 & 85.27 & 89.00 & 84.91 & 87.66 & 84.38 & 89.32 & 83.50 & 82.26 \\
Kimi-K2-thinking & \textbf{91.78} & \underline{82.56} & \textbf{90.40} & \underline{87.05} & \underline{92.46} & \textbf{89.00} & 89.67 & \underline{87.66} & \textbf{91.41} & \underline{86.15} & \underline{84.60} \\
\midrule
\rowcolor{groupgray}
\Block{1-11}{\textbf{Closed-source Models}}& & & & & & & & & & \\
\midrule
Claude-Sonnet-4 & 89.44 & 74.78 & 88.39 & 80.58 & 88.62 & 77.75 & 84.63 & 81.61 & 88.25 & 77.80 & 77.06 \\
Claude-Sonnet-4.5 & 90.22 & 74.00 & 86.83 & 79.69 & 88.36 & 77.49 & 83.88 & 79.60 & 88.05 & 76.97 & 76.79 \\
Claude-Opus-4 & 89.33 & 75.33 & 88.39 & 81.47 & 89.26 & 80.43 & 88.41 & 84.89 & 89.00 & 79.50 & 78.38 \\
Claude-Opus-4.5 & 90.89 & 75.67 & 90.18 & 81.47 & 89.13 & 79.67 & 86.65 & 81.86 & 89.55 & 78.91 & 78.86 \\
GPT-5-chat & 88.11 & 71.11 & 86.61 & 77.23 & 82.74 & 68.67 & 83.12 & 77.33 & 85.40 & 72.42 & 74.02 \\
GPT-5.2-chat & 90.78 & 81.89 & \underline{92.19} & \underline{88.17} & 91.30 & 86.45 & \underline{91.44} & \underline{89.42} & 91.29 & 85.60 & 84.32 \\
\midrule
Claude-Sonnet-4-thinking & 90.78 & 81.00 & 89.73 & 85.94 & 90.28 & 85.29 & 86.65 & 85.39 & 89.79 & 83.89 & 82.75 \\
Claude-Sonnet-4.5-thinking & \textbf{92.44} & 81.00 & 91.07 & \textbf{89.96} & \textbf{92.20} & 88.11 & 90.43 & 88.66 & \textbf{91.81} & 85.99 & 84.45 \\
Claude-Opus-4-thinking & 91.59 & 80.89 & 91.07 & 87.28 & 91.43 & 86.57 & 88.41 & 86.65 & 90.94 & 79.50 & 83.27 \\
Claude-Opus-4.5-thinking & 88.89 & 79.33 & 89.73 & 86.61 & 87.85 & 81.84 & 85.64 & 85.14 & 88.21 & 82.31 & 82.13 \\
Gemini-3.0-pro-thinking & 90.56 & 81.67 & 88.39 & 85.49 & 89.26 & 86.19 & 86.15 & 85.39 & 89.08 & 84.33 & 83.35\\
Gemini-2.5-Pro-thinking & \underline{92.11} & \underline{86.11} & 91.52 & \underline{88.17} & 91.30 & \textbf{90.15} & \textbf{91.94} & \textbf{91.44} & 91.52 & \textbf{88.56} & 86.19\\
GPT-5-thinking & 91.11 & \textbf{86.67} & 90.85 & \textbf{89.96} & 89.64 & 88.62 & 88.16 & 88.16 & 90.15 & \underline{88.09} & \textbf{86.71} \\
GPT-5.2-thinking & \underline{92.11} & 86.00 & \textbf{92.41} & \textbf{89.96} & 91.30 & 88.49 & 89.92 & 89.17 & \underline{91.57} & 87.97 & \underline{86.36} \\
Grok-4-thinking & 91.33 & 85.78 & 89.51 & 87.72 & \underline{91.82} & \underline{89.00} & 90.18 & \underline{89.42} & 90.98 & 87.69 & 85.75\\
Seed-1.6-thinking & 90.00 & 74.78 & 87.72 & 79.91 & 89.13 & 79.80 & 86.65 & 82.87 & 88.80 & 78.51 & 78.50 \\
\midrule
\rowcolor{groupgray}
\Block{1-11}{\textbf{Verifier Model}}& & & & & & & & & & \\
\midrule
General-Verifier & 76.78 & 57.78 & 78.57 & 68.53 & 77.49 & 63.04 & 75.57 & 68.77 & 77.13 & 63.04 & 68.12 \\
Tencent-Qwen2.5-7B-RLVR & 86.22 & 66.33 & 82.14 & 72.77 & 84.65 & 71.23 & 85.14 & 78.34 & 84.84 & 70.87 & 70.17 \\
xVerify-9B-C & 87.11 & \underline{67.44} & 81.92 & 73.44 & 83.89 & 72.55 & 86.39 & 80.10 & 84.92 & 71.98 & 70.91\\
CompassVerifier-32B & \textbf{88.33} & 67.33 & \textbf{88.62} & \underline{77.46} & \textbf{89.13} & \textbf{73.66} & \underline{88.66} & \underline{81.36} & \textbf{88.68} & \underline{73.29} & \underline{73.17}\\
SCI-Verifier-4B & \underline{87.56} & \textbf{68.78} & \underline{87.05} & \textbf{77.68} & \underline{87.98} & \underline{73.53} & \textbf{91.18} & \textbf{84.38} & \underline{88.17} & \textbf{74.28} & \textbf{73.77}\\
\bottomrule
\end{NiceTabular}
}
\captionsetup{singlelinecheck=false, justification=raggedright}
\caption{Performance of different models on \textsc{Prime}. \textbf{F1} refers to the F1-score of \textbf{Overall} field. The best and second-best results are highlighted in \textbf{bold} and \underline{underlined}, respectively.}
\label{table:bench_results}
\end{table*}
In this section, we conduct a comprehensive evaluation on \methodname\ to assess the capability of current models in ensuring process-outcome alignment. We benchmark a diverse array of models, and present an in-depth analysis of performance trends, the impact of reasoning capabilities, and the trade-off between accuracy and efficiency.

\subsection{Experimental Settings.}
\paragraph{Models.}
We categorize the evaluated models into three distinct groups: 
(1) \textbf{Open-source Models}, which encompass standard instruction-tuned models such as Qwen3-Instruct \citep{yang2025qwen3}, DeepSeek-V3.2 \citep{liu2025deepseek}, Phi-4 \citep{abdin2024phi}, and GLM-4.6 \citep{zai2025glm46}, alongside reasoning-enhanced variants like Qwen3-thinking, DeepSeek-V3.2-thinking, and Kimi-K2-thinking \citep{team2025kimi}. This category also includes the GPT-OSS series (20B and 120B) \citep{agarwal2025gpt}.
(2) \textbf{Closed-source Models}, comprising proprietary systems divided into standard chat versions (e.g., GPT-5-chat \citep{openai2025gpt5systemcard}, Claude-Sonnet/Opus \citep{anthropic2025claude4, anthropic2025claude45}) and reasoning-oriented models tailored for complex tasks, such as Gemini-3.0/2.5-Pro-thinking \citep{comanici2025gemini, google2025gemini}, GPT-5-thinking \citep{openai2025gpt5systemcard}, Grok-4-thinking \citep{xai2025grok4}, and Seed-1.6-thinking \citep{bytedance2025seed16}.
(3) \textbf{Specialized Verifiers}, including models explicitly engineered for verification and reward modeling tasks, such as CompassVerifier-32B, xVerify-9B-C, SCI-Verifier-4B, and Tencent-Qwen2.5-7B-RLVR \citep{su-ailabverifier-arxiv-2025}. We also include MathVerify \citep{kydlicek2025mathverify}, a rule-based verifier where we employ GPT-OSS-120B to extract the final answer from the model's response for direct comparison with the ground truth.

\begin{wrapfigure}{r}{0.5\textwidth}
    \centering
    \includegraphics[width=0.95\linewidth]{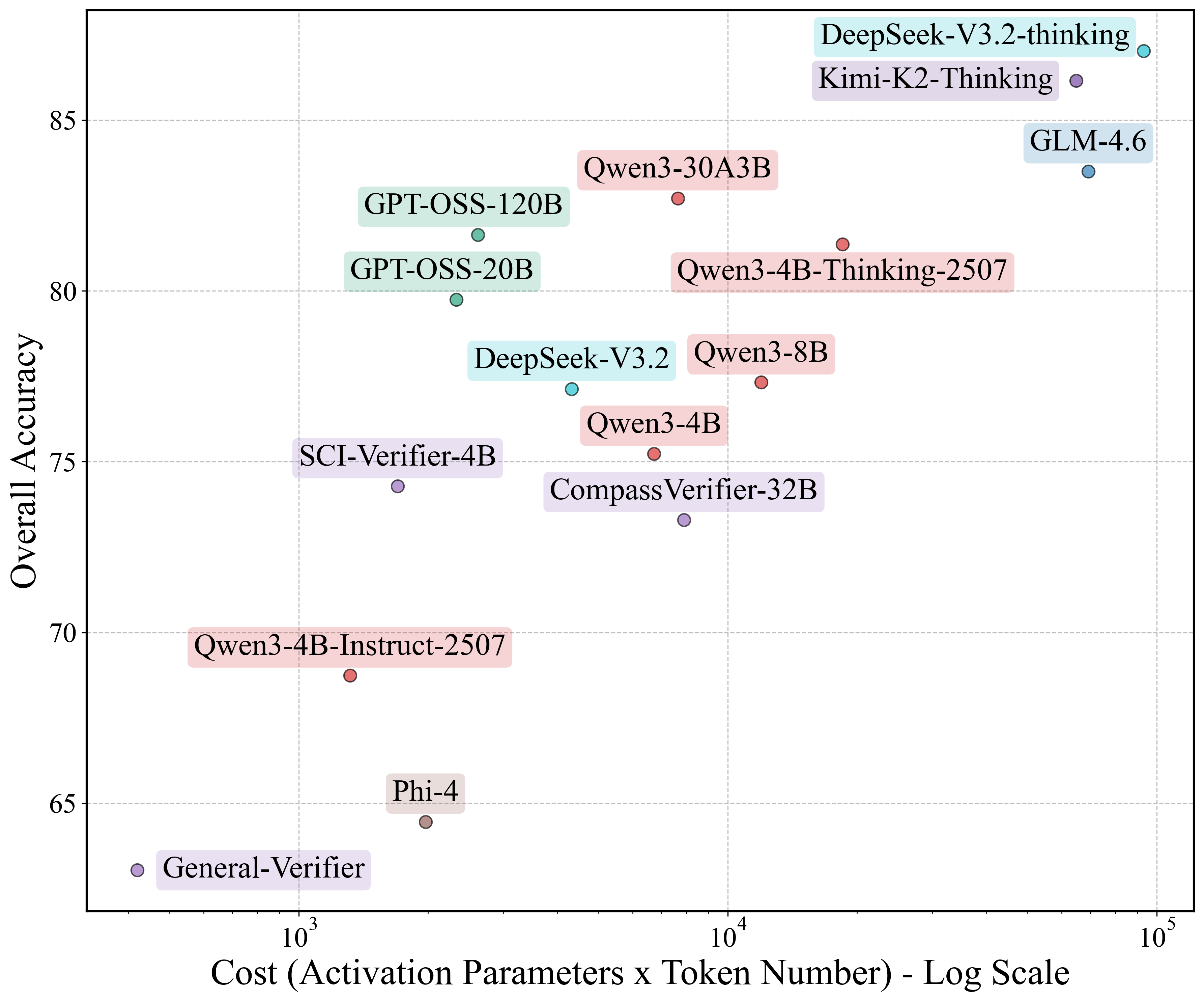}
    \captionsetup{singlelinecheck=false, justification=raggedright}
    \caption{
    Efficiency vs. Performance. Token usage and accuracy comparison. Red dashed line: open-source (below) vs. commercial (above) models.
    }
    \label{fig:f1_vs_tokens}
\end{wrapfigure}

\paragraph{Evaluation Paradigm.}
We adopt distinct evaluation strategies based on the model type. 
For specialized verifiers, we strictly adhere to their official implementations to extract verification predictions, directly comparing them against the corresponding ground truth labels. 
For general-purpose models, we employ a structured prompting approach (Figure~\ref{fig:eval_prompt}) that asks the model to independently check two aspects: whether the derivation logic is sound (\texttt{<process>}) and whether the final answer is correct (\texttt{<outcome>}). 
To determine the model's overall judgment, we simply consider a solution valid only when the model approves \textit{both} the process and the outcome.
Since our benchmark provides specific ground truth labels for both the outcome and the overall correctness, we calculate performance on two corresponding metrics: 
(1) \textbf{Outcome Accuracy}, measured by comparing the model's \texttt{<outcome>} tag directly with the Outcome ground truth $y_{\text{outcome}}$; and 
(2) \textbf{Overall Accuracy}, measured by comparing the model's combined judgment against the Overall ground truth $y_{\text{overall}}$.

\subsection{Performance Analysis.}

\paragraph{General Performance Trends.}
As shown in Table~\ref{table:bench_results}, performances on \methodname are strongly related to both model scale and model family. Closed-source models generally achieve the best results, with Gemini-2.5-Pro-thinking reaching an overall accuracy of 88.56\%. Among open-source models, accuracy increases steadily with parameter count, rising from Qwen3-4B (75.23\%) to GPT-OSS-120B, which attains the best open-source overall accuracy of 81.64\%. These results suggest that large open-source models are becoming increasingly competitive on demanding verification tasks.

\paragraph{Reasoning is Important for Verification.}
Table~\ref{table:bench_results} also shows that models with explicit reasoning consistently outperform their instruction-tuned counterparts. For example, Qwen3-4B-thinking exceeds Qwen3-4B-Instruct by more than 8\%. This gap indicates that verification on \methodname requires more than surface-level semantic matching; it often requires following the derivation to catch logical errors. Overall, strong reasoning ability appears necessary for a high-quality verifier.

\paragraph{Efficiency vs. Performance Trade-off.}
To guide the selection of a verifier for resource-intensive tasks like RLVR, we analyzed the trade-off between computational cost and verification performance. In Figure~\ref{fig:f1_vs_tokens}, we plot the verification accuracy on the \methodname benchmark against a proxy for inference cost (calculated as the product of activated parameters and the number of generated tokens). The visualization reveals a general trend where higher accuracy correlates with an increased proxy cost, indicating that greater reasoning effort often leads to better performance. This analysis provides a practical framework for selecting an appropriate verifier by balancing the required accuracy against the available computational budget for scalable training.

\section{Analysis of Downstream RLVR}
We bridge our \methodname analysis to the practical task of RLVR. To validate the benchmark's utility, we first demonstrate the critical impact of process consistency via an ablation study. We then investigate the correlation between verifier scores and resulting policy performance, establishing \methodname as a strong predictor of downstream RLVR outcomes.

\subsection{Experimental Settings.}

\paragraph{Reinforcement Learning Details.}
 The reward signal is designed to strictly enforce both outcome accuracy and the logical consistency of the derivation. Formally, for a given question $x$, model response $y$, and ground truth answer $a^*$, the reward $R(x, y, a^*)$ is defined as:
\begin{equation}
    R(x, y, a^*) = \mathbb{I}(\mathcal{O}) \cdot \mathbb{I}(\mathcal{C})
\end{equation}
where $\mathbb{I}(\cdot)$ is the indicator function. Here, $\mathcal{O}$ represents the outcome correctness, verifying whether the final answer extracted from $y$ matches the ground truth $a^*$. $\mathcal{C}$ denotes the process consistency, a judgment provided by the verifier on whether the model's reasoning trajectory is logically sound and consistent with the final result. This dual-verification mechanism ensures that the model is penalized for ``lucky guesses'' where the answer is correct but the reasoning is flawed. We perform RLVR directly on two base models Qwen3-8B-Base and Qwen3-14B-Base. The training utilizes the \texttt{WebInstruct-verified} dataset. We configure the training with a global batch size of 64 and a rollout number of 16. All experiments are conducted on a cluster of 40 H800 GPUs, and the models are trained for a total of 300 steps with PPO.

\paragraph{Benchmarks} We evaluate LRMs trained via RLVR using verifiers selected by our benchmark across diverse domains. Our suite includes AIME 2024, AIME 2025, and Beyond-AIME ~\citep{bytedance_seed_2025_beyondaime} for advanced math; GPQA-Diamond (GPQA-D) ~\citep{rein2024gpqa} and SuperGPQA ~\citep{du2025supergpqa} for scientific reasoning; plus MATH-500 and MMLU-Pro ~\citep{wang2024mmlu}. We report average@16 accuracy for the AIME suite, average@4 for GPQA-D and MATH-500, and pass@1 for others.

\begin{table*}[t]
\centering
\resizebox{\textwidth}{!}{
\begin{NiceTabular}{l | c c c c c c c | c}
\toprule
Verifier Model 
& AIME24 
& AIME25 
& Beyond-AIME 
& GPQA-D 
& SuperGPQA 
& MATH-500 
& MMLU-Pro 
& AVG \\
\midrule
\rowcolor{groupgray}
\Block{1-9}{\textbf{Qwen3-8B-Base}}& & & & & & & &\\
\midrule
GPT-OSS-120B-Outcome & 21.04 & 18.65 & 7.94 & 37.18 & 26.91 & 83.20 & 64.90 & 37.12\\
GPT-OSS-120B & \textbf{29.64} & \textbf{22.40} & \textbf{10.94} & \textbf{45.45} & \textbf{31.00} & \textbf{86.40} & \textbf{66.12} & \textbf{41.70}\\
\midrule
\rowcolor{groupgray}
\Block{1-9}{\textbf{Qwen3-14B-Base}}& & & & & & & &\\
\midrule
GPT-OSS-120B-Outcome & 28.85 & 21.61 & 11.27 & 45.17 & \textbf{41.26} & 85.00 & 68.86 & 43.15 \\
GPT-OSS-120B & \textbf{37.14} & \textbf{30.73} & \textbf{18.58} & \textbf{50.66} & 31.60 & \textbf{91.20} & \textbf{69.61} & \textbf{47.07} \\
\bottomrule
\end{NiceTabular}
}
\captionsetup{singlelinecheck=false, justification=raggedright}
\caption{Performance comparison between models trained with Outcome-only verification and our proposed Process-aware verification. The best results are highlighted in \textbf{bold}.}
\label{table:outcome_vs_process}
\end{table*}

\begin{table*}[t]
\centering
\resizebox{\textwidth}{!}{
\begin{NiceTabular}{l | c c c c c c c | c}
\toprule
Verifier Model 
& AIME24 
& AIME25 
& Beyond-AIME 
& GPQA-D 
& SuperGPQA 
& MATH-500 
& MMLU-Pro 
& AVG \\
\midrule
\rowcolor{groupgray}
\Block{1-9}{\textbf{Qwen3-8B Base}}& & & & & & & & \\
\midrule
- & 4.53 & 2.86 & 2.23 & 20.61 & 17.66 & 40.20 & 31.79 & 17.13\\
\midrule
Math-Verify & 15.47 & 14.43 & 7.28 & 40.15 & 26.24 & 78.40 & 63.28 & 35.04\\
CompassVerifier-32B & 13.39 & 10.99 & 7.59 & 38.64 & 30.46 & 78.80 & 59.33 & 34.17\\
Qwen3-4B & 18.65 & 15.78 & 6.92 & 34.38 & 27.01 & 82.80 & 55.24 & 34.40\\
Qwen3-8B & 24.74 & 18.91 & 8.58 & 37.72 & 27.84 & 83.60 & 58.73 & 37.16\\
GPT-OSS-20B & \underline{27.19} & \underline{20.00} & \textbf{11.72} & \textbf{45.93} & \textbf{32.17} & \textbf{86.60} & \underline{65.20} & \underline{41.26}\\
GPT-OSS-120B & \textbf{29.64} & \textbf{22.40} & \underline{10.94} & \underline{45.45} & \underline{31.00} & \underline{86.40} & \textbf{66.12} & \textbf{41.70}\\
\midrule
\rowcolor{groupgray}
\Block{1-9}{\textbf{Qwen3-14B Base}}& & & & & & & \\
\midrule
- & 7.60 & 4.53 & 2.89 & 32.01 & 23.52 & 57.60 & 45.35 & 24.79\\
\midrule
Math-Verify & 20.68 & 18.54 & 7.91 & 47.00 & 28.19 & 84.40 & 66.41 & 39.07\\
CompassVerifier-32B & 21.93 & 13.02 & 8.98 & 46.24 & \textbf{40.50} & 85.00 & 69.08 & 40.68 \\
Qwen3-4B & 28.07 & \underline{21.67} & 10.94 & 46.53 & 35.40 & 85.60 & 67.21 & 42.20\\
Qwen3-8B & 22.40 & 18.75 & 10.92 & 49.21 & 39.03 & 86.40 & 68.27  & 42.14\\
GPT-OSS-20B & \underline{30.36} & 21.51 & \underline{12.92} & \textbf{51.58} & \underline{40.44} & \underline{87.60} & \underline{69.32} & \underline{44.82}\\
GPT-OSS-120B & \textbf{37.14} & \textbf{30.73} & \textbf{18.58} & \underline{50.66} & 31.60 & \textbf{91.20} & \textbf{69.61} & \textbf{47.07} \\
\bottomrule
\end{NiceTabular}
}
\captionsetup{singlelinecheck=false, justification=raggedright}
\caption{Downstream performance comparison. Evaluation of Qwen3-8B Base and Qwen3-14B Base trained via RLVR using different verifiers. The best results are highlighted in \textbf{bold} and the second best are \underline{underlined}.}
\label{table:rl_results}
\end{table*}

\subsection{Impact of Process Consistency.}
To verify the necessity of the process consistency judgment in our reward formulation, we conduct ablation experiments comparing our proposed process-aware RLVR against a standard outcome-only baseline. We use GPT-OSS-120B as the verifier. In the baseline setting, the reward is determined solely by the final answer correctness, ignoring the logical validity of the intermediate reasoning trajectory. The results are presented in Table~\ref{table:outcome_vs_process}. 
We can see that incorporating process-aware supervision yields consistent and significant improvements across both model scales. For the Qwen3-8B-Base model, the inclusion of process verification boosts the average accuracy from 37.12\% to 41.70\%. Similarly, for the Qwen3-14B-Base model, the performance rises from 43.15\% to 47.07\%. The gains are particularly pronounced in high-difficulty reasoning benchmarks; for instance, on AIME 2024, the 8B model improves drastically from 21.04\% to 29.64\%, and the 14B model improves from 28.85\% to 37.14\%. 
This indicates that relying solely on outcome verification makes the policy model prone to reward hacking via ``lucky guesses,'' whereas strictly enforcing process consistency ensures the model learns robust and generalized reasoning patterns, leading to superior performance on complex tasks.

\subsection{Correlation and Performance.}

\paragraph{Verifier Selection for Downstream Validation.}
To assess whether \methodname{} performance reliably predicts utility in practical RLVR pipelines, we select a representative subset of verifiers from Section~\ref{bmk_result}. Instead of relying solely on top-scoring models, we cover a capability spectrum (70\%--85\% Overall Accuracy) by selecting five models: CompassVerifier-32B, Qwen3-4B, Qwen3-8B, GPT-OSS-20B, and GPT-OSS-120B. This diversity allows us to directly correlate benchmark scores with the downstream performance of policy models trained on their reward signals.

\begin{wrapfigure}{r}{0.5\textwidth}
    \centering
    \includegraphics[width=\linewidth]{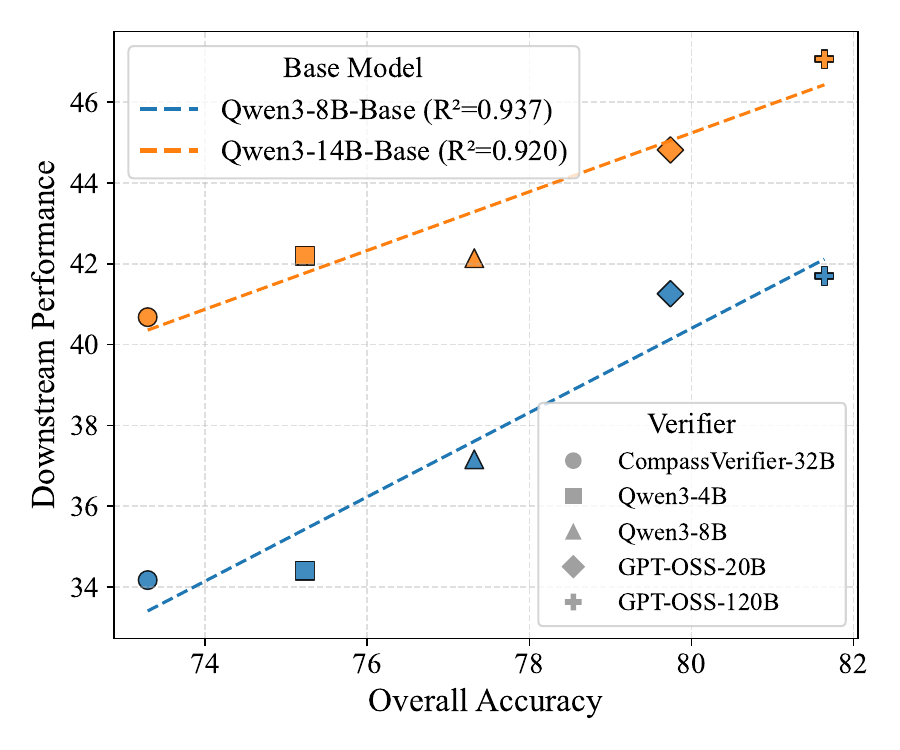}
    \captionsetup{singlelinecheck=false, justification=raggedright}
    \caption{Correlation between verifier performance and downstream policy improvement with verifier. The x-axis represents the Overall Accuracy on our \methodname benchmark, and the y-axis represents the average score on the benchmarks in Table~\ref{table:rl_results}}
    \label{fig:correlation}
\end{wrapfigure}

\paragraph{Correlation between Performance in \methodname and Downstream Performance.}
The comparative results are presented in Table~\ref{table:rl_results} and further visualized in Figure~\ref{fig:correlation}. From Table~\ref{table:rl_results}, we can see that the performance of the trained policy models improves significantly as the capability of the verifier increases. Notably, using our best-performing verifier, GPT-OSS-120B, boosts the average accuracy of the Qwen3-14B Base model from 24.79\% to 47.07\%, surpassing the gains achieved by smaller verifiers like CompassVerifier-32B (40.68\%) or Qwen3-4B (42.20\%). Combining this with the trend analysis in Figure~\ref{fig:correlation}, we observe a strong positive linear correlation between the verifier's Overall Accuracy on \methodname and the policy model's final performance, with coefficient of determination ($R^2$) values reaching 0.937 and 0.920 for the 8B and 14B models, respectively. This indicates that \methodname serves as a highly reliable predictor of a reward model's utility. A verifier that demonstrates high process consistency on our benchmark is statistically more likely to guide the policy model toward correct reasoning paths, thereby effectively preventing reward hacking and enhancing training stability. Furthermore, the training dynamics, visualized in Appendix Figure~\ref{fig:training_curves}, are provided to validate the stability of our RLVR process. The reward curves exhibit a steady increase before reaching a plateau, indicating that each policy model was trained to convergence.

\section{Conclusion}

We introduce \methodname{} to evaluate the process verification capabilities of Large Reasoning Models. Our benchmarking reveals that strong reasoning is a prerequisite for verification, and process-aware verifiers significantly outperform outcome-only baselines. We further validate the benchmark's utility by demonstrating a strong correlation between \methodname{} scores and downstream RLVR policy improvement, confirming that high-quality process supervision leads to substantial gains in complex reasoning tasks.

\section*{Limitation}

A primary limitation of this study is that we did not train a specialized process-aware verifier model from scratch. Instead, we leveraged existing high-performance open-source model selected by our benchmark to serve as the verifier in RLVR. Our empirical findings indicate that these general-purpose strong reasoners already possess sufficient capability to distinguish between valid reasoning and spurious ``lucky guesses,'' thereby providing highly effective supervision signals without the need for costly specialized training. Future work will focus on scaling this verification paradigm to even larger LRMs and extending the evaluation to agentic scenarios, where multi-step planning and tool use require even more process consistency checks.

\section*{Ethical Considerations}
The development of \textsc{PRIME} adheres to strict ethical standards for AI research. Our initial corpus of 7 million problems is sourced from university-level textbooks and public examinations. This data is intended solely for academic purposes to evaluate reasoning capabilities, and the final 2,530 benchmark samples are the result of an original, multi-stage processing and annotation pipeline. Professional domain experts in STEM performed all fine-grained labeling and were compensated fairly for their professional expertise. To protect privacy, no personally identifiable information of these experts is included in the released dataset. A primary goal of this work is to mitigate reward hacking and ``lucky guesses'' in RLVR by enforcing strict process-outcome alignment. By penalizing pseudo successes, we promote the development of more trustworthy and logically sound reasoning models for scientific applications. While we release the benchmark and trained models to facilitate open research, they are not intended for autonomous decision-making in high-stakes environments without human oversight. We do not foresee significant dual-use risks, as the content is centered on fundamental mathematics and engineering disciplines.

\newpage
\bibliography{main}

\newpage
\appendix
\appendixpage  

\appendix
\section{Appendix}
\label{sec:appendix}

\begin{figure}[h]
    \centering
    \includegraphics[width=0.5\textwidth]{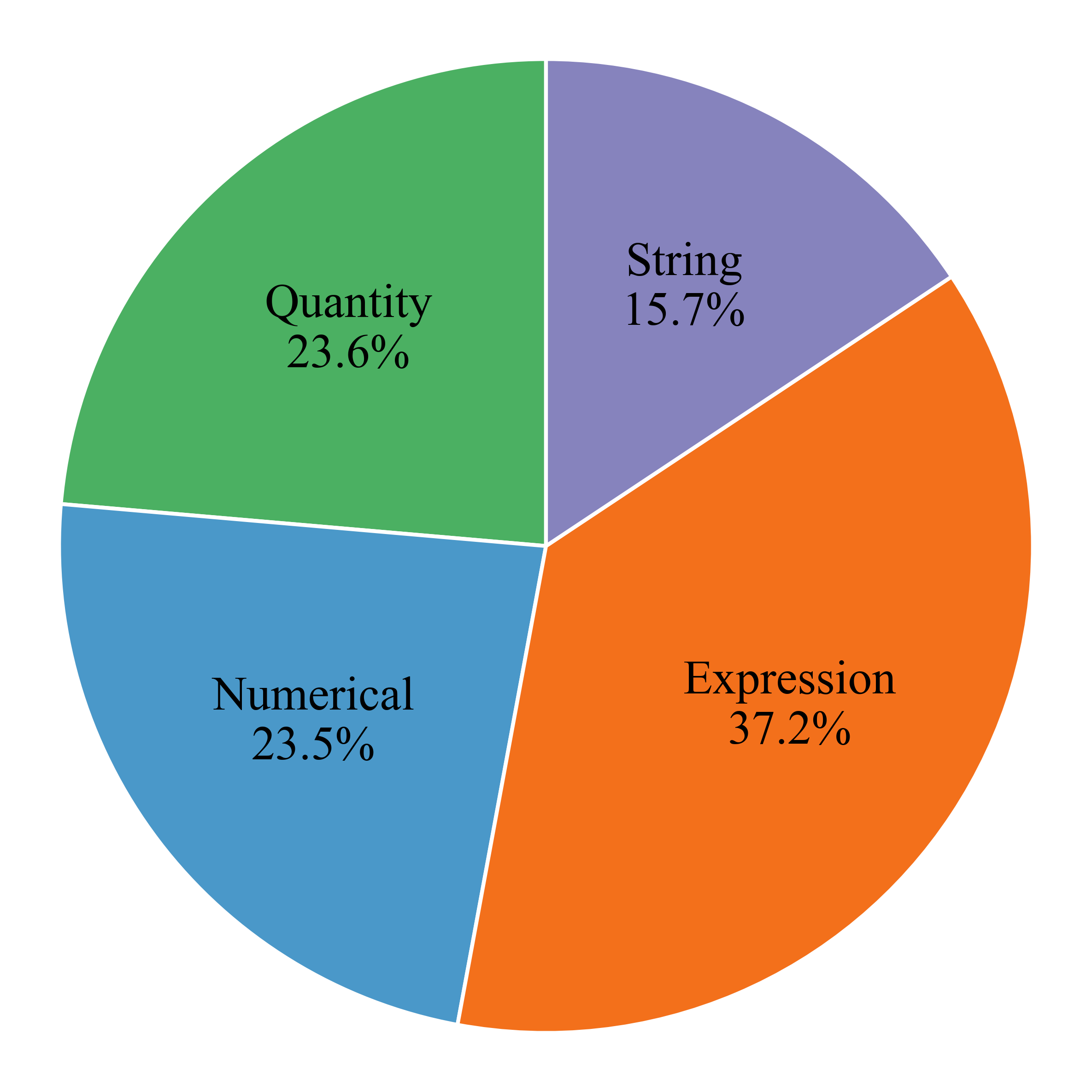}
    \caption{\textbf{Answer type distribution of \methodname.}}
    \label{fig:answer_type_dist}
\end{figure}

\begin{figure*}[t]
    \centering
    \includegraphics[width=\textwidth]{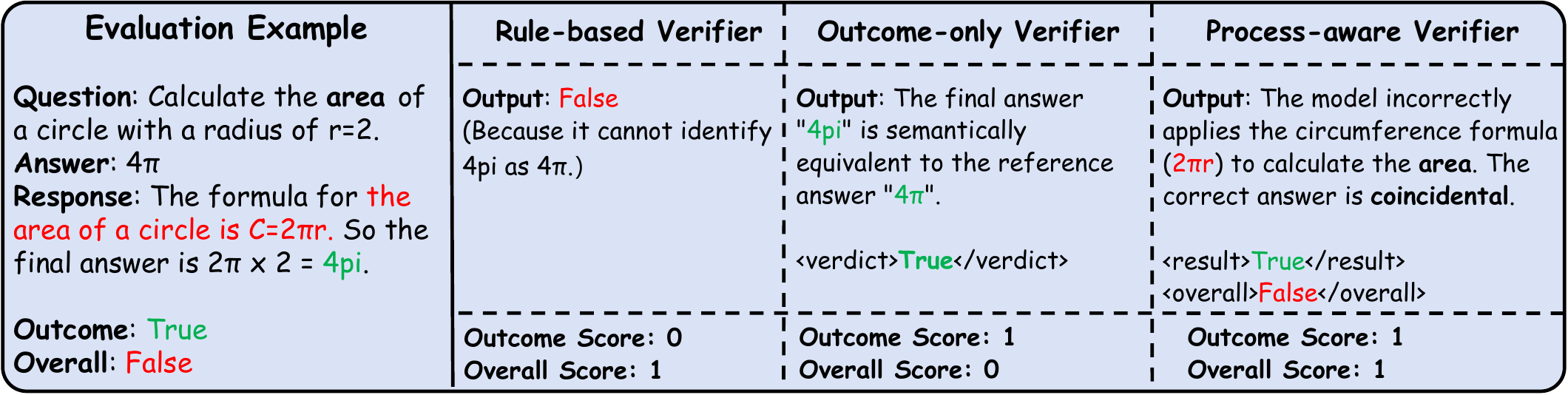}
    \caption{Qualitative comparison of verification paradigms.}
    \label{fig:eval_example}
\end{figure*}

\begin{figure*}[t]
    \centering
    \begin{subfigure}[b]{0.48\textwidth}
        \centering
        \includegraphics[width=\textwidth]{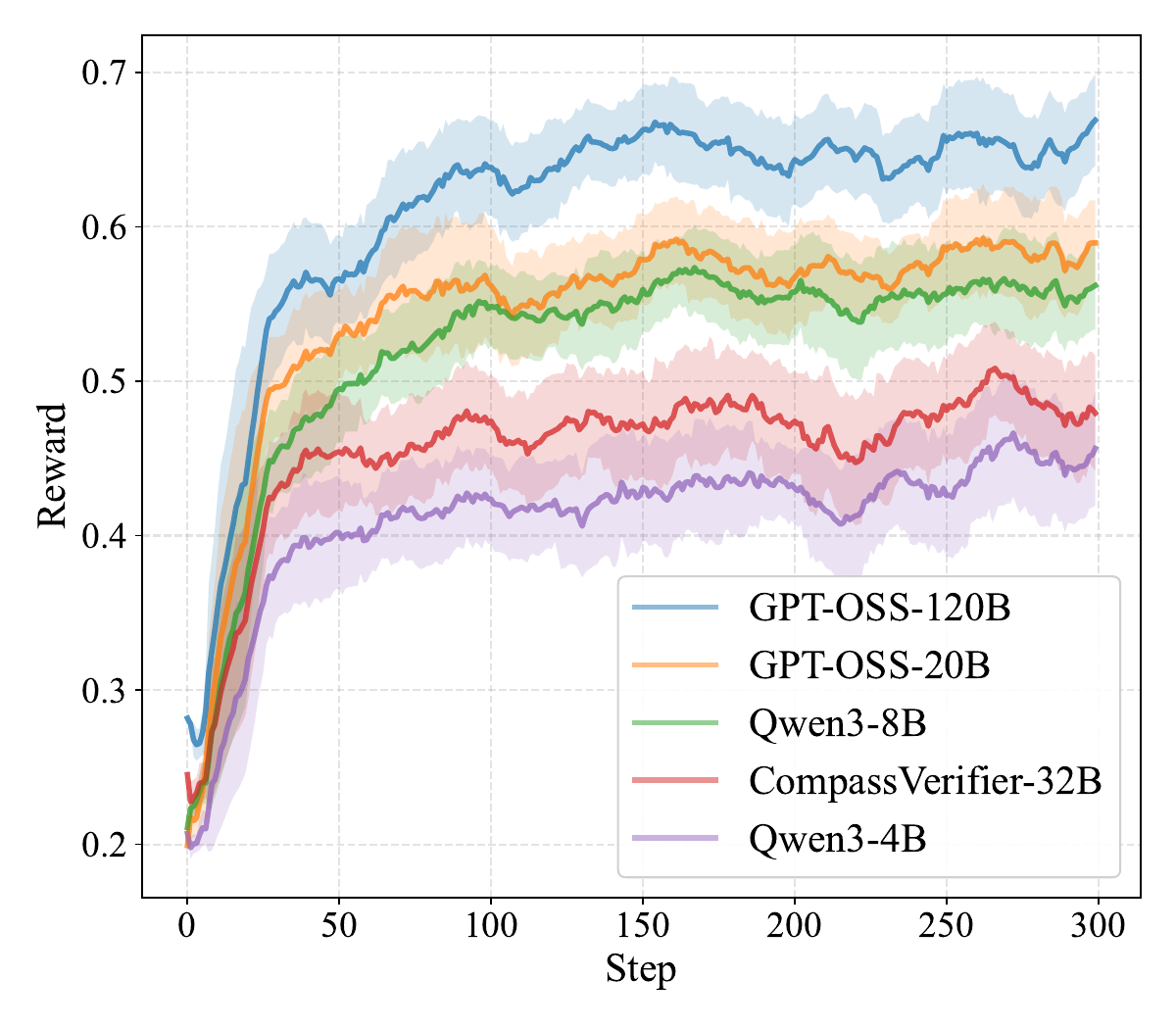}
        \caption{Qwen3-8B-Base}
        \label{fig:8b_curve}
    \end{subfigure}
    \hfill
    \begin{subfigure}[b]{0.48\textwidth}
        \centering
        \includegraphics[width=\textwidth]{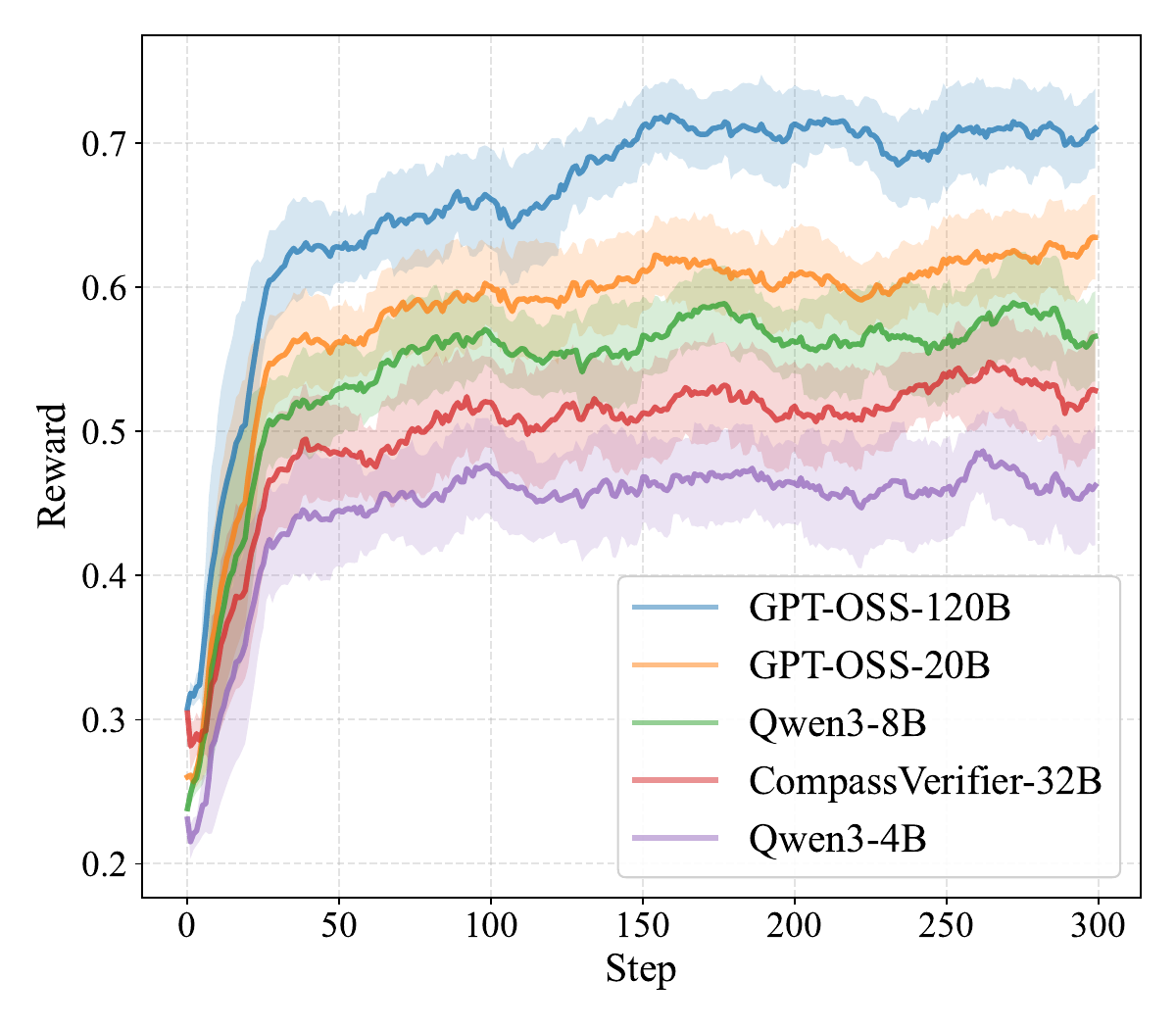}
        \caption{Qwen3-14B-Base}
        \label{fig:14b_curve}
    \end{subfigure}
    \captionsetup{singlelinecheck=false, justification=raggedright}
    \caption{RLVR training reward curves using different verifiers. Higher-performing verifiers identified by \methodname consistently provide more stable and higher reward signals during the training process.}
    \label{fig:training_curves}
\end{figure*}

\begin{table}[h]
\centering
\begin{NiceTabular}{l c}
\toprule
Model & Accuracy \\
\midrule
CompassVerifier & 4.66 \\
xVerify-9B-C & 11.89 \\
Tencent-Qwen-7B-RLVR & 8.86 \\
Sci-Verifier & 9.09 \\
Qwen3-4B & 22.14 \\
Qwen3-8B & 29.37 \\
GPT-OSS-20B & 36.83 \\
GPT-OSS-120B & 45.92 \\
\bottomrule
\end{NiceTabular}
\caption{Performance comparison on the ``Lucky Guess'' subset in our \methodname. This subset contains samples with correct final outcomes but flawed derivations, measuring the verifiers' sensitivity to derivation false positives.}
\label{table:luckyguess}
\end{table}

\begin{figure*}[t]
\centering
\begin{tcolorbox}[
    colback=red!5!white,          
    colframe=magenta!60!black,    
    coltitle=white,               
    title=\textbf{The Evaluation Prompt for Process-Outcome Consistency Verification},
    fonttitle=\small\bfseries,
    fontupper=\scriptsize\ttfamily\linespread{0.8}\selectfont,
    sharp corners,    
    boxrule=0.8pt,
    enhanced,
    breakable
]

\# Role

You are an extremely meticulous and conscientious exam grader. Your task is to assess whether a student’s submitted answer is correct.

\# Input Variables

- **[Question]**: The problem that needs to be solved.

- **[Reference Answer]**: Typically only the final result, as a reference.

- **[Student Answer]**: Includes the full solution steps and the final result.

\# Primary Principles

I will provide you with a question and the corresponding reference answer. You need to determine:

1. Whether the student’s solution **process** is correct.

2. Whether the student’s final **result** is correct.

3. Whether the student’s answer is **perfect**.

\# Core Logic
\#\# Process Correctness

You must carefully check whether there are problems in each step of the student’s solution.
Your judgment should follow the principle of “presumption of correctness”: the process is considered correct unless one of the following fundamental errors occurs.

**Fundamental Errors**:

1. Logical errors, calculation errors, or factual errors in reasoning.

2. Inconsistencies within reasoning, such as inconsistency between statements.

**Note**: If the student did not provide the solution process, it is considered **correct by default**.

\#\# Result Correctness

Similarly, your judgment of the final result also follows the principle of “presumption of correctness”: the answer is considered correct unless one of the following fundamental errors occurs.

**Fundamental Errors**:

1. The student fails to provide a clear final answer.

2. The student’s final answer does not satisfy the problem requirements.

3. The student’s final result does not match the reference answer (unless you can prove the student’s result is also correct).

**Note**: In numerical calculations, discrepancies due to reasonable rounding are **not considered** fundamental errors but should be carefully distinguished from computational errors.

\#\# Answer Perfection

An answer is considered perfect only when both the process and result are correct, and none of the following imperfections occur. Any answer with a fundamental error is necessarily imperfect.

Imperfection reference standards:

1. Insufficient precision: The solution does not provide the exact form (e.g., $\sqrt{2}$) but only an approximate value (e.g., 1.414).

2. Not simplified: Fractions, radicals, etc., are not expressed in simplest form.

3. Redundant content: Contains information clearly irrelevant to the solution.

4. Lack of process: Missing necessary solution steps.

5. Other flaws you consider imperfect but not fundamental errors.

**Note**: If the problem’s requirements conflict with the standards of perfection, the problem requirements take precedence.

\# Special Cases

- If the student’s answer contains a large amount of repetitive content, both the process and the result should be considered incorrect.

- If the student’s answer is incomplete (e.g., cut off), both the process and the result should be considered incorrect.

\# Output Format

You need to output a result in XML format with the following fields:

<process> [Your judgment on the correctness of the solution process, only True or False] </process>
<outcome> [Your judgment on the correctness of the final result, only True or False] </outcome>
<perfect> [Your judgment on whether the answer is perfect, only True or False] </perfect>
<reason> [Your reasoning for the judgment in a few brief sentences] </reason>\\
\{few\_shots\}\\
\# Task

Now, please strictly follow the above principles, logic, and format to evaluate the following input and return the result in XML format.

<question>
{question}
</question>

<student\_answer>
{student\_answer}
</student\_answer>

<reference\_answer>
{reference\_answer}
</reference\_answer>
\end{tcolorbox}
\vspace{-1.em}
\caption{Detailed evaluation prompt used for assessing verifier performance on \methodname. This prompt enforces a multi-dimensional check on process correctness, result accuracy, and overall perfection.}
\label{fig:eval_prompt}
\end{figure*}

\section{Case Study}
Figure~\ref{fig:eval_example} illustrates a ``lucky guess'' scenario where the model incorrectly uses the circumference formula ($C=2\pi r$) to calculate the area of a circle ($r=2$) yet coincidentally arrives at the correct value $4\pi$. This case highlights the limitations of baseline verifiers. The rule-based verifier rejects the response due to rigid formatting constraints as it fails to match ``4pi'' with ``$4\pi$'', resulting in a false negative regarding the outcome. In contrast, the outcome-only verifier accepts the response based solely on the correct final answer and ignores the flawed derivation. This generates false positive signals that encourage reward hacking in RLVR. Finally, a process-aware verifier (such as GPT-OSS-120B) successfully detects the logical fallacy of applying the circumference formula for area calculation and correctly penalizes the response. It further indicates that process-aware verification is essential to filter out false positive responses, ensuring that rewards are assigned solely for reasoning that is both accurate in outcome and valid in derivation.

\setcounter{figure}{0}
\makeatletter
\renewcommand{\thefigure}{A\@arabic\c@figure}
\makeatother

\setcounter{table}{0}
\makeatletter
\renewcommand{\thetable}{A\@arabic\c@table}
\makeatother

\end{document}